\documentclass[11pt]{article}

\usepackage[margin=1in]{geometry}
\usepackage[authoryear,round]{natbib}

\usepackage[utf8]{inputenc}
\usepackage[T1]{fontenc}
\usepackage[hidelinks]{hyperref}
\usepackage{url}
\usepackage{booktabs}
\usepackage{amsfonts}
\usepackage{nicefrac}
\usepackage{microtype}
\usepackage{xcolor}

\usepackage{amsmath,amssymb,amsthm,mathrsfs,mathtools,bm}
\usepackage{graphicx}
\usepackage{subcaption}
\usepackage{placeins}

\newcommand{\suozhi}[1]{}
\newcommand{\lh}[1]{}

\ifdefined\hideauthorcomments
    \renewcommand{\suozhi}[1]{}
\fi

\title{GraphCast--Mamba: Learning Non-Markovian Corrections for Weather Forecasting}

\author{
\normalsize Ilya Vilkoviskiy$^{1,*}$, Liang-Hong Mo$^{1,*}$, Suozhi Huang$^{2}$, Dmitry A. Abanin$^{1,3,4,5}$\\[0.5em]
\small $^{1}$Department of Physics, Princeton University, Princeton, NJ 08544, USA\\
\small $^{2}$Department of Electrical and Computer Engineering, Princeton University,\\
\small Princeton, NJ 08544, USA\\
\small $^{3}$\'Ecole Polytechnique F\'ed\'erale de Lausanne (EPFL), 1015 Lausanne, Switzerland\\
\small $^{4}$Princeton Quantum Initiative, Princeton University, Princeton, NJ 08544, USA\\
\small $^{5}$Google Research, Brandschenkestrasse 150, 8002 Z\"urich, Switzerland
}
\date{}

\begin{document}

\maketitle
\begingroup
\renewcommand{\thefootnote}{{*}}
\footnotetext[1]{{\textit{Equal contribution}}}
\endgroup

\begin{abstract}
Unresolved atmospheric processes can introduce history dependence beyond the short input windows used by many weather forecasting models. We introduce GraphCast--Mamba, which augments a frozen, pretrained \(1^\circ\) GraphCast backbone with a trainable Mamba correction branch combining graph message passing and Mamba layers. The branch retains a temporal memory state across successive six-hour forecast steps, learning corrections informed by temporal context beyond GraphCast's two-frame input. Training is confined to the correction branch, using ERA5 data from 2015--2021. After 2,500 optimizer updates, including warm-up, the selected model achieves substantial improvements relative to the uncorrected GraphCast backbone. {It reduces the {GraphCast loss (normalized weighted mean-squared error; MSE)} by 14.42\% when aggregated over the full ten-day forecast and by 21.66\% at day 10 (240\,h after forecast initialization).} {At this final forecast lead time, root-mean-square error (RMSE) decreases by 7.09--14.60\% across all 11 evaluated variable groups, spanning temperature, wind, pressure, geopotential, specific humidity, and precipitation.} These results demonstrate broad improvements in ten-day weather forecasts through a short correction-training schedule while retaining the pretrained backbone.

\end{abstract}

\section{Introduction}

{Learned global weather models produce skillful forecasts from one or a few recent atmospheric states~\citep{bi2023pangu,lam2023graphcast,chen2023fuxi}. GraphCast, for example, uses two states six hours apart to predict the next state. These inputs provide only a partial description of the atmosphere: they contain a limited set of variables at finite spatial resolution, while unresolved processes can still influence their evolution. The history of the resolved state may carry information about these influences that a short input window does not capture.}

{This is a familiar closure problem in turbulent-flow modeling. Resolving the smallest dynamically active scales is computationally costly~\citep{yang2021grid}, so practical models must account for their effects on the resolved scales. A common approach uses approximate dissipative closures, such as eddy viscosity or hyperviscosity, to represent transfer to unresolved scales. These closures can reproduce selected aspects of the flow, but their accuracy depends on the quantity and scale of interest. For example, in forced two-dimensional turbulence, viscous, hyperviscous, and Leith closures can approximate enstrophy transfer, whereas the Smagorinsky closure dissipates enstrophy at the wrong scales~\citep{pietarila2013closures}. Resolution studies provide a related caution: mean dissipation and enstrophy can be nearly converged while higher-order moments of velocity gradients, dissipation, and enstrophy, together with distribution tails, remain substantially underestimated~\citep{DonzisResolutionEffects}. Reproducing bulk statistics therefore does not ensure accurate fluctuations or extreme events.}

{A complementary approach allows the closure to depend on the history of the resolved state. Unresolved variables have their own dynamics, so their influence need not be determined by the current resolved variables alone; past observations can contain information about those hidden degrees of freedom. The Mori--Zwanzig formalism makes this explicit: eliminating unresolved variables generally yields an instantaneous term, a history-dependent memory term, and a term associated with unresolved initial conditions, even when the full dynamics are Markovian~\citep{YenTingLinMZ2021,de2026data}. Retaining the memory contribution leads to a \emph{non-Markovian} closure, in which the resolved evolution depends on past states as well as the present one. This motivates persistent temporal memory as a mechanism for representing unresolved influences that an instantaneous dissipative approximation may miss.}

{Evidence from learned partial differential equation (PDE) models supports this motivation. Memory Neural Operator (MemNO), which combines a neural operator with a structured state-space sequence (S4) layer, improves predictions from coarse or noisy observations~\citep{buitrago2025memory}. Here, we ask whether retaining temporal context beyond GraphCast's two-frame input improves forecast corrections when the correction architecture and parameter count are held fixed.}

{To investigate this question, we introduce GraphCast--Mamba, which augments a frozen pretrained $1^\circ$ GraphCast backbone with a trainable correction branch combining graph message passing and Mamba layers~\citep{lam2023graphcast,gu2023mamba}. Graph message passing exchanges spatial information, while Mamba retains temporal context across successive six-hour forecast steps. The branch adds a history-dependent correction to each backbone forecast, and the corrected forecast supplies the next input. Only the correction branch is trained.}

{Our principal experiment trains the correction on ERA5 data from 2015--2021, selects the model on 32 forecast initializations from the 2022 development set, and evaluates ten-day forecasts on 122 initializations in 2023 at $1^\circ$ resolution.} This experiment tests the benefit of correction for one pretrained backbone at a fixed resolution. Its dependence on resolution and transfer to other forecasters remain open questions.

Our contributions are threefold.
\begin{itemize}
    \item \textbf{A temporal-memory correction model.} GraphCast--Mamba combines spatial message passing with temporal Mamba layers in a separate Mamba correction branch, extending the predictor's temporal context while keeping the pretrained GraphCast backbone frozen.
    \item \textbf{Evidence for a contribution from temporal memory.} A matched control trained and evaluated without persistent temporal memory uses the same architecture, parameter count, and optimization schedule. Retaining memory improves ten-day GraphCast loss reduction by $5.26$ percentage points (pointwise paired 95\% confidence interval: $4.70$--$5.86$) for this training realization.
    \item {\textbf{Broad improvements in ten-day forecasts.} After 2,500 optimizer updates, including warm-up, our model reduces the {GraphCast loss} by $14.42\%$ aggregated over the ten-day rollout and by $21.66\%$ at the final 240-hour lead.} {At that final lead, RMSE decreases by $7.09$--$14.60\%$ across all 11 evaluated variable groups relative to the paired frozen backbone.}
\end{itemize}

\begingroup
\section{Related Work}
\label{sec:related_work}

\begin{figure}[!tp]
    \centering
    \begin{subfigure}{\linewidth}
        \includegraphics[width=0.93\linewidth]{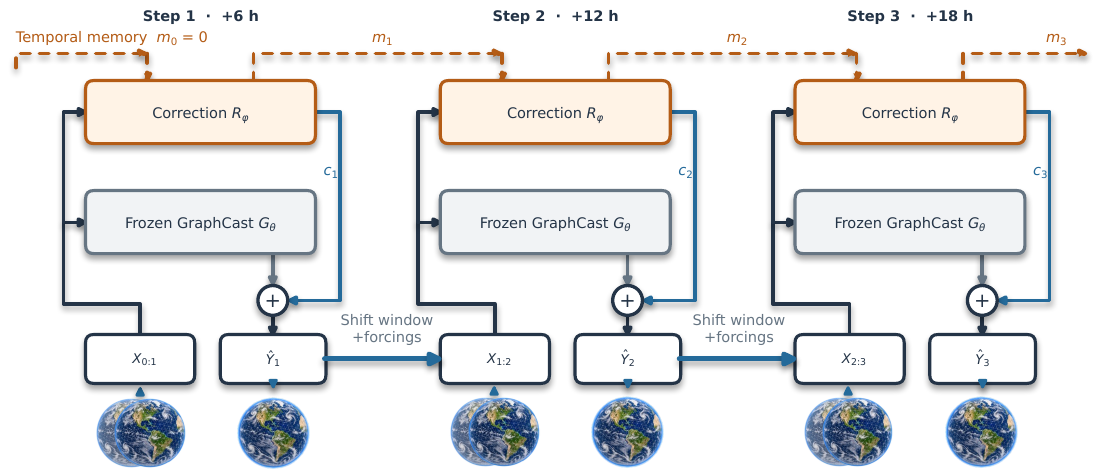}
        \caption{Autoregressive rollout over three six-hour steps. {Each GraphCast--Mamba forecast advances the two-frame input window; dashed orange arrows carry the temporal memory state}, initialized here with $m_0=0$.}
        \label{fig:GC_Mamba_rollout}
    \end{subfigure}
    \par\smallskip
    \begin{subfigure}{\linewidth}
        \includegraphics[width=0.93\linewidth]{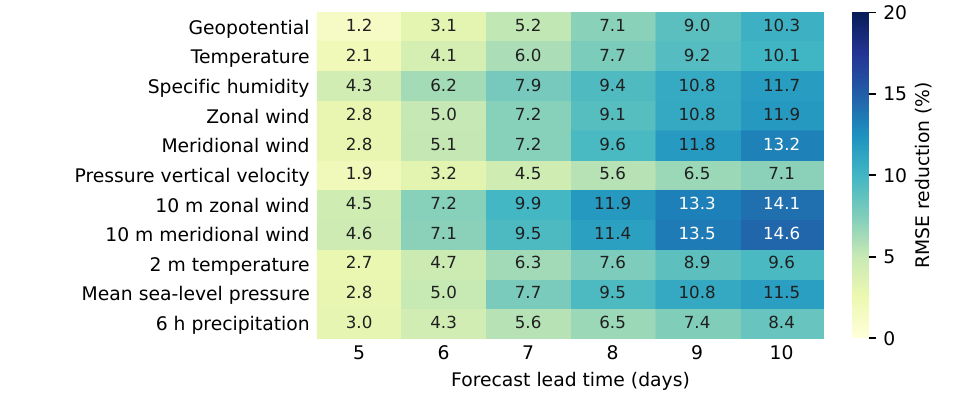}
        \caption{{Variable-group RMSE reductions at days 5 through 10.}}
        \label{fig:breadth_7yr}
    \end{subfigure}
    \par\smallskip
    \begin{subfigure}{\linewidth}
        \includegraphics[width=0.93\linewidth]{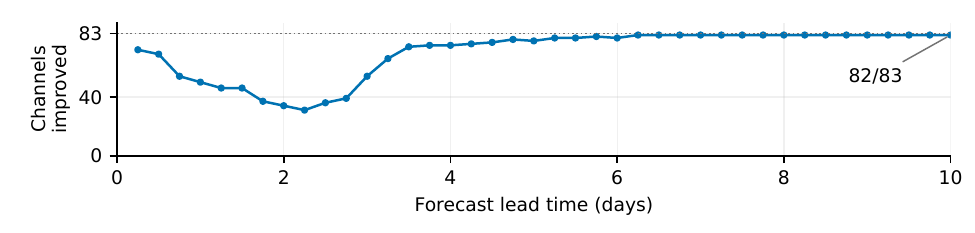}
        \caption{{Channel scores: channels with lower RMSE than GraphCast at each six-hour lead.}}
        \label{fig:coverage_7yr}
    \end{subfigure}
    \caption{{GraphCast--Mamba: temporal-memory correction and forecast-skill overview. Panel (a) shows forecast feedback and memory carry. In (b), positive percentages indicate lower variable-group RMSE than the paired GraphCast baseline; (c) counts channels with lower RMSE, out of 83. Counts describe aggregate differences without channel-wise significance tests. Section~\ref{sec:eval_protocol} defines the evaluation protocol and RMSE aggregation.}}
    \label{fig:GC_Mamba}
\end{figure}

\begin{figure}[!htbp]
    \centering
    \includegraphics[width=\linewidth]{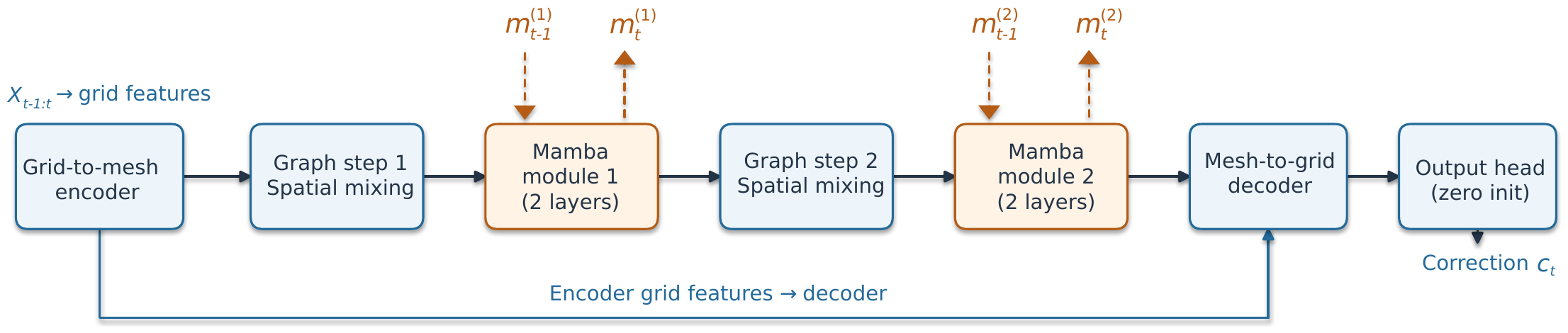}
    \caption{{Trainable Mamba correction branch.} Two graph message-passing steps each feed a two-layer Mamba module, followed by decoding and an output head initialized to zero. {The frozen GraphCast backbone forecast $G_\theta(X_t)$ is combined with the learned forecast correction $c_t$ to produce $\hat{Y}_t$. The Mamba correction branch has separate parameters and a temporal memory state $m_t$}, comprising the SSM states and causal-convolution caches of its four Mamba layers.}
    \label{fig:GC_Mamba_correction}
    \vspace{-10pt}
\end{figure}

Our work connects three lines of research: memory-based modeling of partially observed dynamics, temporal memory in weather forecasting, and learned corrections to existing predictors.

\paragraph{Memory and learned closures in reduced dynamics.}
When only part of a dynamical system is resolved, the evolution of the retained variables can depend on their history, as described by the Mori--Zwanzig formalism~\citep{YenTingLinMZ2021}. Motivated by this connection, MemNO combines Fourier neural operators with S4 memory for coarse or noisy PDE observations~\citep{buitrago2025memory}; SS-NO combines spatial state-space operators with temporal memory~\citep{koren2025ssno}. Memory also enters learned corrections to imperfect dynamics: \citet{gupta2021neuralclosures} augment reduced models with neural delay differential equations, while \citet{levine2022modelerror} develop a framework for learning both memoryless and memory-dependent model error. These studies motivate using temporal context to represent effects missing from a predictor's resolved state. Our experiments assess the predictive value of retained context without identifying a physical memory kernel.

\paragraph{Temporal memory in weather forecasting.}
Recurrent weather models already retain information across forecast steps. SwinRNN carries hidden features through global forecasts, while SwinVRNN adds learned perturbations for ensemble prediction~\citep{hu2023swinvrnn}; SwinRDM couples recurrent forecasts with diffusion super-resolution~\citep{chen2023swinrdm}. For radar nowcasting, GauMamba combines Mamba processing with gated recurrent unit (GRU) memory to predict evolving Gaussian representations~\citep{wang2025gaumamba}. These models establish temporal memory as a forecasting mechanism. Our architecture places memory in a separate correction branch: Mamba processes time at each mesh node, graph message passing exchanges spatial information, and the pretrained GraphCast backbone remains frozen.

\paragraph{Learned forecast corrections.}
Learning corrections provides a way to improve an existing predictor while retaining its forecast model. $\delta$-Adapter learns input adjustments and output residual corrections around frozen time-series forecasters~\citep{liang2026forecast}. In numerical weather prediction, \citet{chen2022correcting} learn state-dependent corrections from analysis increments and apply them during FV3-GFS integration. Solver-in-the-Loop trains corrections through differentiable solver rollouts, accounting for their effects on subsequent states~\citep{um2020solver}. These approaches establish both frozen-model adaptation and correction during model integration. For our weather predictor, the remaining question is whether a correction benefits from temporal context beyond the backbone's two-frame input.

These precedents motivate our investigation of persistent temporal memory in a correction branch for a frozen pretrained global weather model. Our experiments compare corrections with and without retained state using the same architecture and parameter count.
\par
\endgroup

\section{Method: A hybrid GraphCast--Mamba architecture}

We study whether {a temporal memory state improves forecasts beyond the finite input history supplied to the frozen GraphCast backbone}. GraphCast uses two successive atmospheric states as input; its forecast is Markovian with respect to this augmented state. Figure~\ref{fig:GC_Mamba_rollout} shows {how the Mamba correction branch corrects each GraphCast backbone forecast and carries its temporal memory state across successive six-hour steps. The GraphCast--Mamba forecast advances the meteorological input window, while the temporal memory state retains information from the preceding trajectory.} Figure~\ref{fig:GC_Mamba_correction} details how graph message passing mixes information across mesh nodes and Mamba updates memory independently at each node. This introduces history dependence relative to the baseline input representation.

\subsection{{Architecture}}

\paragraph{{GraphCast backbone}}
{The frozen GraphCast backbone uses a pretrained checkpoint and also serves as the standalone forecast baseline.} This graph neural network operates on a latitude--longitude grid and an auxiliary mesh, with graph $\mathcal{G}(\mathcal{V}^{\textrm{G}},\mathcal{V}^{\textrm{M}},\mathcal{E}^{\textrm{G2M}},\mathcal{E}^{\textrm{M}},\mathcal{E}^{\textrm{M2G}})$. The vertices $\mathcal{V}^{\textrm{G}}$ represent the meteorological grid and $\mathcal{V}^{\textrm{M}}$ the mesh. Encoding uses grid-to-mesh edges $\mathcal{E}^{\textrm{G2M}}$, followed by $N=16$ graph message-passing steps over $\mathcal{E}^{\textrm{M}}$ and decoding over mesh-to-grid edges $\mathcal{E}^{\textrm{M2G}}$. Latent node features have width $W_{\textrm{GC}}=512$. The mesh provides a more compact representation for spatial processing than the meteorological grid.

\paragraph{{Mamba correction branch: temporal memory on the mesh.}}
The Mamba correction branch uses Mamba~\citep{gu2023mamba} to process forecast time independently at each mesh node, with spatial exchange handled by graph message passing. Each layer normalizes the width-$512$ mesh features, projects them into a $d_{\mathrm{inner}}$-dimensional temporal representation, and projects its output back to the mesh width through a residual connection. We use $d_{\mathrm{inner}}=16$. Across autoregressive forecast calls, each layer retains both its {state-space model (SSM) state} and its three-token causal-convolution cache separately for each batch element and mesh node.

Our model uses an SSM state dimension of $16$ per inner channel, a causal convolution width of $4$, a rank-$32$ projection for the step-size parameter $\Delta_t$, and one shared group of input-dependent $B_t,C_t$ coefficients. Additional studies examine $d_{\mathrm{inner}}=32$. The memory-reset controls in Section~\ref{sec:memory_ablation} use $d_{\mathrm{inner}}=16$ and one shared $B,C$ group to test the contribution of temporal memory.

\paragraph{An additive correction with temporal memory.}
Let $X_t$ denote the two-frame meteorological input (shown as $X_{t-1:t}$ in the diagrams) and the associated forcings. Forecast calls are indexed from $t=1$; $\hat{Y}_t$ denotes the next-state forecast produced by call $t$. The temporal memory state $m_t$ collects the SSM states and causal-convolution caches of all Mamba layers. As shown in Figure~\ref{fig:GC_Mamba_rollout}, {GraphCast--Mamba combines the frozen GraphCast backbone $G_\theta$ with the trainable Mamba correction branch $R_\phi$, whose output $c_t$ is the forecast correction}:
\vspace{-4pt}
\begin{equation}
    (c_t,m_t)=R_\phi(X_t,m_{t-1}),
    \qquad
    \hat{Y}_t=G_\theta(X_t)+c_t,
    \qquad \theta\ \text{fixed}.
    \label{eq:additive_correction}
\end{equation}
{The Mamba correction branch in Figure~\ref{fig:GC_Mamba_correction}} has its own grid-to-mesh encoder, mesh processor, and mesh-to-grid decoder. It receives {the same meteorological inputs as the GraphCast backbone} and uses separate parameters throughout. Its mesh processor uses two graph message-passing steps, each followed by a two-layer Mamba module, giving four recurrent layers in total. Encoder grid features are also passed directly to the decoder. Suppressing this grid-feature connection in the notation, the mesh-processing path is
\begin{align}
    L_t^{(0)} &= \operatorname{Encode}_\phi(X_t),\\
    \widetilde{L}_t^{(k)} &= \operatorname{GNN}^{(k)}_\phi(L_t^{(k-1)}),\\
    (L_t^{(k)},m_t^{(k)})
    &=\operatorname{MambaModule}^{(k)}_\phi
       (\widetilde{L}_t^{(k)},m_{t-1}^{(k)}),\qquad k=1,2,\\
    c_t &= \operatorname{Decode}_\phi(L_t^{(2)}).
\end{align}
{Each Mamba module contains two residual Mamba layers; $m_t^{(k)}$ comprises their SSM states and causal-convolution caches.} The output head is initialized to zero, so {the initial GraphCast--Mamba forecast equals the GraphCast backbone forecast}. During autoregressive forecasting, the oldest input frame is dropped, {the GraphCast--Mamba forecast $\hat{Y}_t$ is appended}, and the known forcings are updated to form $X_{t+1}$. {The temporal memory state is passed separately to the next correction-branch call.} The cold-start rollout illustrated in Figure~\ref{fig:GC_Mamba_rollout} begins with $m_0=0$ and {retains the updated temporal memory state thereafter}.

\subsection{Training}
\label{sec:training}
\paragraph{Initialization and trainable parameters.}
The Mamba correction branch inherits pretrained spatial weights and uses a zero-initialized output head, so the initial GraphCast--Mamba forecast matches the GraphCast backbone forecast. We train all correction-branch parameters $\phi$ on ERA5 2015--2021 while keeping the separate GraphCast backbone parameters $\theta$ frozen.

\begin{figure}[!htbp]
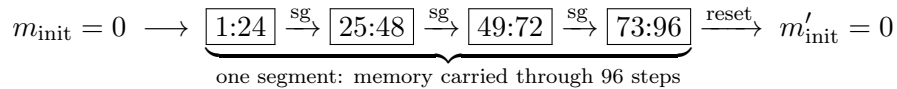

    \centering
    \normalsize
    \[
    m_{\mathrm{init}}=0\;\longrightarrow\;
    \underbrace{
      \boxed{1{:}24}
      \xrightarrow{\mathrm{sg}}\boxed{25{:}48}
      \xrightarrow{\mathrm{sg}}\boxed{49{:}72}
      \xrightarrow{\mathrm{sg}}\boxed{73{:}96}
    }_{\text{one segment: memory carried through 96 steps}}
    \xrightarrow{\mathrm{reset}}\;m_{\mathrm{init}}'=0
    \]
    \caption{Memory carry during training. Boxes show the four 24-step windows of one segment. The arrows marked $\mathrm{sg}$ carry the SSM states and causal-convolution caches unchanged across window boundaries while stopping their gradients. Both are reset for the next segment.}
    \label{fig:training_segments}
\end{figure}

\paragraph{Training segments and memory carry.}
We divide the training sequence into contiguous segments of 96 six-hour steps (24 days), each processed as four consecutive 24-step windows (Figure~\ref{fig:training_segments}). The temporal memory state, comprising both the SSM states and causal-convolution caches, starts at zero for each segment and is carried through all 96 steps. Each window begins with a four-step teacher-forced prefix (ERA5 inputs), followed by 20 feedback-conditioned forecast steps that use the GraphCast--Mamba forecast as feedback. Meteorological inputs therefore return to ERA5 at the start of each window, while the correction branch retains the preceding window's temporal memory state. Gradients propagate through the temporal memory state within a window; at its end, the state is carried forward with its gradient detached. Gradients are also stopped through the meteorological feedback.

\paragraph{Training loss.}
At each step, the forecast correction is supervised against the residual target $r_t=Y_t-G_\theta(X_t)$, where $Y_t$ is ERA5 truth and $X_t$ is the current input. Since $c_t-r_t=\hat{Y}_t-Y_t$, this minimizes the {GraphCast loss} $\ell_{\mathrm{GC}}$ of the GraphCast--Mamba forecast. We average the loss over all $H=24$ steps, including the teacher-forced prefix, and the $B$ batch elements:
\begin{equation}
    \mathcal{J}(\phi)
    =\frac{1}{BH}\sum_{b=1}^{B}\sum_{k=1}^{H}
      \ell_{\mathrm{GC}}(\hat{Y}_{b,k},Y_{b,k}).
    \label{eq:training_objective}
\end{equation}
Appendix~\ref{app:training_loss} gives the full loss definition and weights.

\paragraph{Training schedule.}
We train with AdamW for 2,000 post-warm-up updates after a 500-update warm-up, or 2,500 optimizer updates in total. The evaluated model uses stochastic weight averaging (SWA); Appendix~\ref{app:training} provides the complete training configuration, model-selection procedure, and computational costs.

\section{Results}
\label{sec:numerical}

We train the GraphCast–Mamba correction branch on seven years of ERA5 data (2015–2021), keeping the GraphCast backbone frozen. We then evaluate whether the model improves forecast skill, how broadly the gains extend across the atmospheric state, and how strongly performance depends on temporal memory. To assess the latter, we compare against a matched reset-trained control and a memory-reset evaluation of the selected checkpoint.

\subsection{Evaluation protocol and metrics}
\label{sec:eval_protocol}

\begin{figure}[!t]
    \centering
    \includegraphics[width=\linewidth]{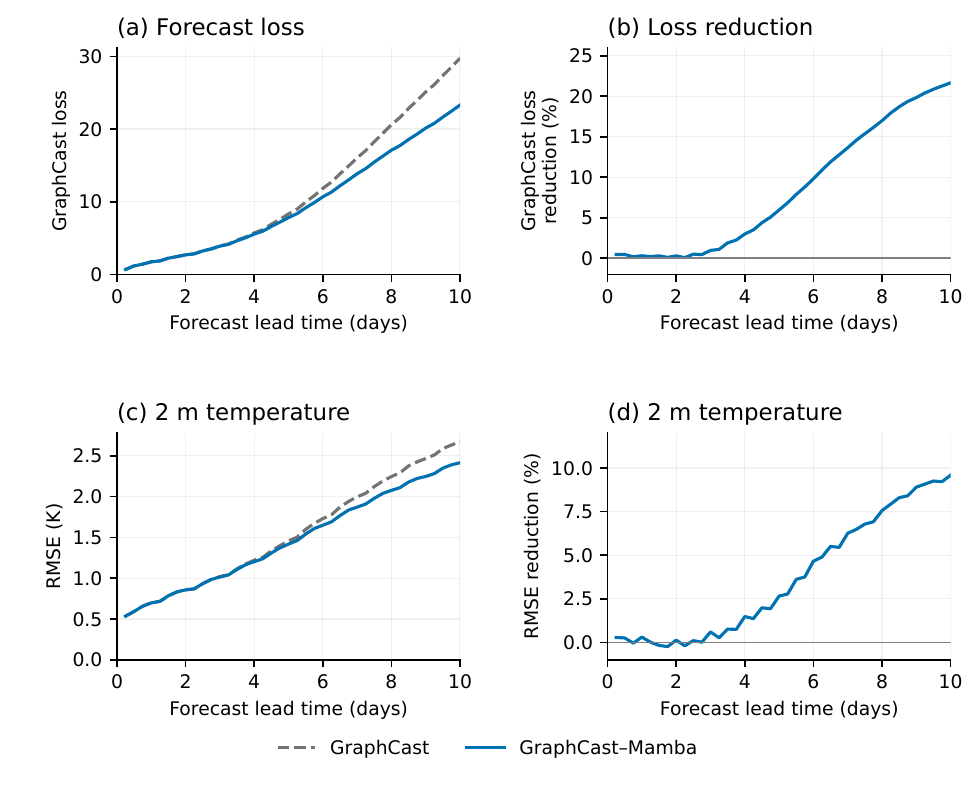}
    \caption{{GraphCast--Mamba forecast skill on the 122 evaluation initializations in 2023 (Section~\ref{sec:eval_protocol}). Top: {GraphCast loss} and its per-lead reduction. Bottom: $2$\,m temperature RMSE and RMSE reduction on the same forecast initializations. The Mamba correction branch uses $d_{\mathrm{inner}}=16$, a threefold Mamba learning rate (LR), and SWA over post-warm-up updates 500--2,000. Every reduction uses the paired GraphCast baseline shown in the absolute-error panels.}}
    \label{fig:res1_forecast}
\end{figure}

The principal experiments use GraphCast--Mamba with the original pretrained $1^\circ$ GraphCast backbone and the Mamba correction branch trained in Section~\ref{sec:training}. The same checkpoint supplies the lead-time, variable-group, and channel-coverage results. {The principal results use 122 forecast initializations at 00\,UTC every third day from 1 January to 30 December 2023, hereafter the evaluation initializations, on the native $1^\circ$ grid at 40 six-hour forecast steps from 6 to 240\,h.} {The checkpoint was selected among the six learning-rate arms on 32 forecast initializations from the 2022 development set (Appendix~\ref{sec:training_sensitivity}) and fixed for this 2023 evaluation.} 

Each forecast takes two ERA5 states as input. GraphCast--Mamba initializes the temporal memory state to zero, and advances for ten days using its own predictions as feedback. GraphCast--Mamba feeds back its full prediction, while GraphCast follows an independent uncorrected trajectory. Stateful models retain their updated temporal memory state between steps; the controls without persistent temporal memory reset it at every step.

We first examine aggregate forecast performance using the {GraphCast loss}, with difference-standard-deviation normalization, latitude weights, pressure-level weights, and variable weights. For forecast initializations indexed by $s$, let $\mathcal{L}_{s,k}$ denote this loss at six-hour lead step $k$. The ten-day rollout loss reduction is
\begin{equation}
    \mathcal{I}_{1:40}=100\left(1-
    \frac{\sum_s\sum_{k=1}^{40}\mathcal{L}^{\mathrm{model}}_{s,k}}
         {\sum_s\sum_{k=1}^{40}\mathcal{L}^{\mathrm{GC}}_{s,k}}\right).
    \label{eq:v3_loss_reduction}
\end{equation}
Day-10 loss corresponds to $k=40$ only. Losses are aggregated before taking the ratio; the rollout score is not an average of per-lead percentages.

For each variable and lead, RMSE pools squared physical-unit errors across forecast initializations using normalized cosine-latitude weights before taking the square root. Upper-air variable-group RMSE additionally pools squared errors equally over 13 pressure levels; channel RMSE scores each surface variable and each upper-air pressure level separately. We report $(1-\mathrm{RMSE}_{\mathrm{model}}/\mathrm{RMSE}_{\mathrm{GC}})\times 100\%$, with positive values denoting lower error.

\subsection{Forecast skill improves at longer forecast lead times}

Figure~\ref{fig:res1_forecast} shows that the benefit of the correction is concentrated at longer forecast lead times. {Relative to the GraphCast baseline, the {GraphCast loss} decreases by $14.42\%$ over the full ten-day rollout and by $21.66\%$ at the final 240-hour lead.}

Physical-unit errors show a similar pattern. {For $2$\,m temperature, RMSE changes remain small through day 3, before the reduction grows at longer forecast lead times.} {At day 10, RMSE falls from $2.67$\,K to $2.41$\,K, a $9.59\%$ reduction.}

\subsection{Gains extend across the predicted atmospheric state}
The gains at longer forecast lead times extend across the predicted atmospheric state. {All 11 variable groups have lower RMSE at every evaluated six-hour lead from day 5 through day 10, with day-10 reductions of $7.09$--$14.60\%$ (Figure~\ref{fig:breadth_7yr}).} These improvements span circulation, thermodynamic, moisture, and surface quantities. Table~\ref{tab:variable_group_rmse} in Appendix~\ref{app:variable_group_rmse} reports the reductions by variable at selected leads, together with absolute day-10 RMSEs.

{Appendix~\ref{app:city_traces} complements these aggregate results with individual forecast trajectories at Paris and New York City, including both improvements and larger local errors in a historical January 2022 forecast.}

\subsection{\texorpdfstring{{Comparison with higher-resolution GraphCast}}{Comparison with higher-resolution GraphCast}}
\label{sec:common_grid}
\begin{figure}[!htbp]
    \centering
    \includegraphics[width=\linewidth]{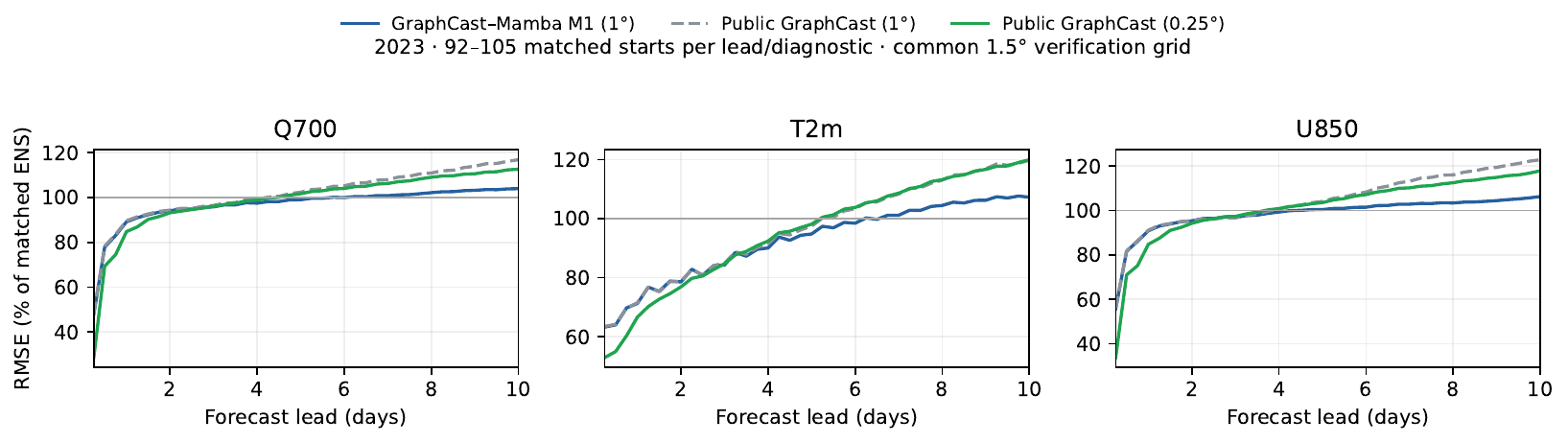}
    \caption{Forecast RMSE relative to the matched ECMWF ensemble mean (ENS) on a common $1.5^\circ$ grid. From left to right: 700\,hPa specific humidity (Q700), $2$\,m temperature (T2m), and 850\,hPa zonal wind (U850). Blue shows the selected GraphCast--Mamba model M1 ($1^\circ$), dashed gray its public GraphCast\_small backbone ($1^\circ$), and green the public GraphCast checkpoint ($0.25^\circ$). From 122 candidate 00\,UTC initializations every third day in 2023, each diagnostic and lead uses the identical ENS-available dates for all four forecasts (92--105 starts per point in these panels). Squared errors are pooled before taking square roots. The horizontal line marks ENS at $100\%$; lower values indicate lower RMSE. Appendix~\ref{app:common_grid} specifies the protocol.}
    \label{fig:common_grid_selected}
\end{figure}

Figure~\ref{fig:common_grid_selected} compares the selected $1^\circ$ GraphCast--Mamba model with its frozen GraphCast\_small backbone and the public $0.25^\circ$ GraphCast checkpoint~\citep{lam2023graphcast}. Forecasts are initialized throughout 2023 and verified against identical ERA5 fields on a common $1.5^\circ$ grid, with RMSE expressed relative to the matched ECMWF ensemble mean (ENS). At each diagnostic and lead, every predictor uses the same available dates. GraphCast--Mamba has lower RMSE than the $0.25^\circ$ checkpoint for all 12 diagnostics at every evaluated lead from day 3.5 through day 10 (Appendix~\ref{app:common_grid}). At day 10, reductions are $7.70\%$ for 700\,hPa specific humidity, $10.58\%$ for $2$\,m temperature, and $9.83\%$ for 850\,hPa zonal wind, spanning $7.70$--$11.42\%$ across the 12 diagnostics.

Despite operating at $1^\circ$ resolution, GraphCast--Mamba achieves better long-lead accuracy than the more computationally demanding $0.25^\circ$ GraphCast model on the common verification grid. The reference uses approximately 16 times as many horizontal grid points, four times as many mesh vertices, and 37 rather than 13 pressure levels, increasing spatial computation and activation-memory requirements. This comparison concerns the specified fixed models on the common verification grid; it does not isolate the effect of resolution or measure skill at native $0.25^\circ$ scales. ENS retains lower RMSE on all 12 diagnostics at day 10.
\suozhi{\color{blue}I think it is interesting that our GC-Mamba outperforms even the much larger model.}

\subsection{{Temporal memory contributes to forecast skill}}
\label{sec:memory_ablation}
The preceding results show that GraphCast--Mamba improves forecasts at longer lead times across the predicted atmospheric state. The correction branch introduces both additional trainable capacity and temporal memory. {To examine the contribution of temporal memory, we compare the selected model with a matched control trained and evaluated with memory reset at every step. We also reset the selected model's memory only during forecasting to measure its dependence on retained state.}

{Table~\ref{tab:memory_controls} summarizes three policies obtained from two training runs. Policy names specify memory handling during training first and forecasting second. Carry passes the updated SSM state and causal-convolution cache to the next step; reset zeros both before every step. Resetting leaves all learned weights and the two-frame meteorological input in place: the full correction branch still computes a correction, but cannot retain its temporal memory state between calls. Every training segment and independent forecast begins with a zero temporal memory state under all three policies.}

\begin{table}[!htb]
    \centering
    \small
    \begin{tabular}{lcc}
        \toprule
        Policy & \multicolumn{2}{c}{Temporal memory retained} \\
        \cmidrule(lr){2-3}
        Training/forecasting & Training & Forecasting \\
        \midrule
        Carry/carry & $\checkmark$ & $\checkmark$ \\
        Carry/reset & $\checkmark$ & $\times$ \\
        Reset/reset & $\times$ & $\times$ \\
        \bottomrule
    \end{tabular}
    \caption{{Memory retained between successive steps: $\checkmark$ denotes retention; $\times$ denotes resetting both the SSM state and causal-convolution cache before every step. Carry/carry and carry/reset share the same trained weights. Reset/reset is trained separately with the same architecture and parameter count.}}
    \label{tab:memory_controls}
\end{table}

{Carry/carry and carry/reset use the selected checkpoint. Reset/reset is trained independently with memory reset in both its 500-update warm-up and 2,000-update main stage. Both training configurations use seven years of data, a $K=20$ feedback-conditioned training tail, inner dimension $d_{\mathrm{inner}}=16$, and the same batch size, optimizer schedule, training seed, and SWA checkpoint indices (Appendix~\ref{app:training_configuration}).} We evaluate all three policies on the same 122 forecast initializations and 40 lead times of Section~\ref{sec:eval_protocol}, using a common GraphCast baseline. {Appendices~\ref{app:training_loss} and~\ref{app:memory_evaluation} describe the loss calculation and paired uncertainty estimates.}

\begin{figure}[!htb]
    \centering
    \includegraphics[width=0.8\linewidth]{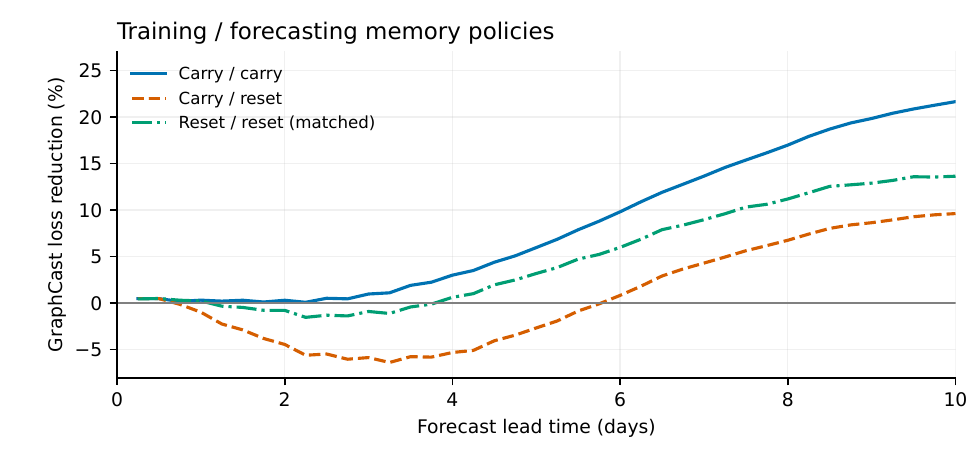}
    \caption{{{\textbf{Comparison of carry/carry, carry/reset, and matched reset/reset memory policies defined in Table~\ref{tab:memory_controls}.} Curves show exact GraphCast loss reduction relative to a common GraphCast baseline on the 122 initializations in 2023. Reset/reset uses the matched training schedule described in the text.}}}
    \label{fig:memory_reset}
\end{figure}

{Over the full ten-day rollout, carry/carry achieves a $14.42\%$ GraphCast loss reduction, compared with $9.16\%$ for matched reset/reset. The paired advantage is $5.26$ percentage points (pointwise 95\% confidence interval: $4.70$--$5.86$). At day 10, the reductions are $21.66\%$ and $13.64\%$, respectively, a difference of $8.02$ percentage points (pointwise 95\% confidence interval: $7.20$--$9.05$). Figure~\ref{fig:memory_reset} shows all three policies.}

{Resetting the selected model's memory only during forecasting (carry/reset) yields $4.57\%$ rollout and $9.62\%$ day-10 loss reduction. Its rollout reduction is $9.85$ percentage points below carry/carry.} This shows that the model uses its temporal memory state, although the drop may partly reflect the mismatch between training and forecasting. {The matched training-policy comparison supports a benefit from persistent memory for this configuration and training realization. The intervals quantify forecast-sample uncertainty for fixed checkpoints; variation across training seeds is not measured.}

\section{Conclusion}

Our results show that a small correction branch with persistent temporal memory can improve forecasts from a frozen pretrained weather model at longer forecast lead times. {The matched reset-trained control supports a benefit from retaining temporal memory for the tested configuration and training realization, while inference-time reset demonstrates the selected model's dependence on retained history.} This suggests that compact temporal extensions offer a promising way to improve existing forecasters with limited additional training. How this benefit depends on spatial resolution and the choice of backbone remains an open question.

{\textbf{Limitations.} Our results show that non-Markovian memory can improve forecasts from a frozen Markovian backbone. This suggests the possibility of jointly learning the forecasting dynamics and temporal memory within a native model. However, autoregressive training introduces sequential dependencies between forecast steps, limiting temporal parallelism and opportunities for batching and data reuse. Backpropagation through long rollouts also requires storing or recomputing intermediate activations, making efficient computation and memory access central challenges for joint training of the full model. To keep training tractable, our study restricts learning to a correction branch attached to a frozen pretrained forecaster.

A separate direction concerns high spatial frequencies. At fixed resolution, dynamically important unresolved scales may increase the value of temporal history, as suggested by the PDE studies discussed in the introduction. Regimes with substantial high-frequency content could therefore provide a more sensitive probe of non-Markovian memory. Our aggregate loss and RMSE metrics do not establish whether memory improves fine-scale predictions or reduces excessive smoothing; this requires evaluating scale-dependent forecast errors and spatial-gradient statistics.}
\FloatBarrier

\clearpage
\section*{AI use statement}

We used a large language model for minor editorial assistance, including correcting grammar and
typo and refining formatting. It did not contribute to the development of research ideas, the analysis,
or the claims.

\section*{Ethics Statement}
All datasets, models in this work are publicly available. The usage of these datasets, models, code strictly follows the terms and conditions set by the providers. 

\section*{Reproducibility Statement}
To ensure reproducibility of our results, we provide detailed documentation of our experimental setup and methodology in the main paper and appendix. The code for data, evaluation,
and training will be open-sourced upon publication.

\section*{{Acknowledgments}}
{Ilya Vilkoviskiy and Dmitry Abanin acknowledge support through a Brown Investigator Award and grant 864597 from the European Research Council (ERC), funded by the European Union's Horizon 2020 research and innovation programme. Ilya Vilkoviskiy also acknowledges support from OpenAI through its ChatGPT for Academic Researchers program~\citep{openai2026academicresearchers}.}

\bibliographystyle{plainnat}
\bibliography{bib,arxiv_references}

\clearpage
\appendix

\section{\texorpdfstring{{Training details and model selection}}{Training details and model selection}}
\label{app:training}

\subsection{Training configuration}
\label{app:training_configuration}
The configuration below supplies the principal {GraphCast loss} and RMSE results in Section~\ref{sec:numerical}.

\paragraph{Backbone, data, and normalization.}
We use the public \texttt{GraphCast\_small} checkpoint trained on ERA5 1979--2015, with $1^\circ$ resolution, 13 pressure levels, a mesh refined through level 5, and precipitation as both an input and an output. Its filename is {\small\texttt{GraphCast\_small - ERA5 1979-2015 - resolution 1.0 - pressure levels 13 - mesh 2to5 - precipitation input and output.npz}}.

Training examples are drawn from ERA5 2015--2021, with 2022 reserved for development. These data come from the six-hourly WeatherBench2 ERA5 archive,\footnote{\path{gs://weatherbench2/datasets/era5/1959-2023_01_10-wb13-6h-1440x721_with_derived_variables.zarr}.} subsampled by selecting every fourth latitude and longitude of the $0.25^\circ$ grid to obtain a $181\times360$ grid at $1^\circ$. We retain pressure levels 50, 100, 150, 200, 250, 300, 400, 500, 600, 700, 850, 925, and 1000\,hPa. We reuse the GraphCast normalization files \path{mean_by_level.nc}, \path{stddev_by_level.nc}, and \path{diffs_stddev_by_level.nc}, distributed under \path{gs://dm_graphcast/graphcast/stats/}, throughout correction training and evaluation.

\paragraph{Initialization and batching.}
We initialize the Mamba correction branch from the frozen GraphCast backbone checkpoint: its grid-to-mesh encoder (G2M), mesh-to-grid decoder (M2G), and two graph message-passing steps copy the corresponding pretrained weights, using the first two of GraphCast's 16 graph message-passing steps. We insert newly initialized Mamba modules after these graph message-passing steps and append a linear output head with weights and biases initialized to zero. Consequently, the initial forecast correction is zero and the GraphCast--Mamba forecast equals the pretrained GraphCast backbone forecast. The copied spatial parameters, Mamba modules, and output head form the trainable parameters $\phi$; the separate GraphCast backbone parameters $\theta$ remain frozen throughout training on ERA5 2015--2021. Each optimizer update processes one window from each of four independent segments, giving batch size $B=4$ under the memory-carry procedure in Section~\ref{sec:training}.

\paragraph{Optimizer and checkpoint settings.}

The reported model uses two optimization stages. The first stage, called the 500-update warm-up in the main text, linearly increases the learning rate from zero to $10^{-4}$ over its first 50 updates and then holds it constant. Spatial and Mamba parameters use the same learning rate during this stage. The second stage transfers the learned correction weights and starts a fresh optimizer for 2,000 further updates, resetting the random-number generator, temporal memory, and data cursor. Its spatial learning rate ramps from zero to $10^{-4}$ over 50 updates and then follows cosine decay toward $10^{-5}$ over the remaining 1,950 updates. The Mamba learning rate is three times the spatial learning rate throughout this second stage.

Both stages use AdamW with moments $(0.9,0.98)$, weight decay $10^{-4}$ applied to all trainable correction parameters, global gradient-norm clipping at 1, and random seed 22. Model computation uses bfloat16, with float32 meteorological fields and temporal-state boundaries. The configuration uses spatial width 512, inner dimension 16, and one $B,C$ group. The evaluated model averages the parameters saved at second-stage updates 500, 1,000, 1,500, and 2,000 using SWA; these are the post-warm-up checkpoint indices used elsewhere in the paper.

\paragraph{Matched reset-training control.}

The reset/reset control uses the same backbone, training data, correction architecture, parameter count, batch size four, random seed 22, precision, optimizer schedules, and SWA indices as the selected carry-trained model. Both its 500-update warm-up and 2,000-update main stage reset the SSM states and causal-convolution caches before every correction prediction; the main stage starts from its own reset-trained warm-up weights. The 24-step training windows, $K=20$ feedback-conditioned tail, full correction feedback, and loss objective are retained. Hyperparameters are fixed to the selected configuration without further model selection. The control uses the recovered archived training and model code, with previously unarchived dependencies supplied from the workspace and frozen for the new run. The historical environment was not fully archived, so matching refers to the configured training protocol and does not imply bitwise reproduction of the earlier run.

\subsection{Complete loss definition}
\label{app:training_loss}
The per-step loss in Equation~\ref{eq:training_objective} is the {GraphCast loss}:
\begin{equation}
    \ell_{\mathrm{GC}}(\hat{Y},Y)
    =\sum_v\frac{\lambda_v}{N_{\mathrm{g}}L_v}
      \sum_{i=1}^{N_{\mathrm{g}}}\sum_{\ell=1}^{L_v}
      a_i q_{v,\ell}
      \left(\frac{\hat{Y}_{v,\ell,i}-Y_{v,\ell,i}}
      {\sigma_{v,\ell}}\right)^2.
    \label{eq:training_gc_loss}
\end{equation}
Here $N_{\mathrm{g}}$ is the number of grid points and $L_v$ is the number of levels for variable $v$. The scales $\sigma_{v,\ell}$ are GraphCast's standard deviations of six-hour differences. The latitude weights $a_i$ have unit mean and include the pole-cell correction. For upper-air variables, $q_{v,\ell}=p_\ell/\overline{p}$ weights levels in proportion to pressure; surface variables have $L_v=1$ and $q_{v,1}=1$. The variable weights $\lambda_v$ are 1 for upper-air variables and $2$\,m temperature, and 0.1 for the two $10$\,m wind components, mean sea-level pressure, and six-hour accumulated precipitation.

\paragraph{{Evaluation variants.}}
We use GraphCast loss as the common label for the weighted MSE diagnostics. {The principal results and updated 2023 memory comparisons evaluate Equation~\ref{eq:training_gc_loss} directly, using a common GraphCast baseline.} {The closed-loop training-tail sweep reconstructs it from channel RMSEs using per-level normalization and pressure weights, without the pole-cell correction.} The archived open-loop sweep instead uses variable-group RMSEs and level-averaged normalization scales. These reconstructions therefore approximate the loss above.

\subsection{Memory-control evaluation and uncertainty}
\label{app:memory_evaluation}

The matched reset/reset evaluation reuses the original frozen evaluator, initialization manifest, input data, and normalization statistics. Carry/carry and carry/reset reuse their original forecast files. All three policies start with zero temporal memory, feed back their full corrected forecasts, and are scored on the same 122 initializations and 40 six-hour leads; late-December forecasts use verifying ERA5 states through 9 January 2024. All loss reductions use the original carry/carry evaluation's uncorrected GraphCast trajectory as their common denominator, pooling losses before computing ratios.

The carry/carry versus matched reset/reset rollout contrast is a separate planned analysis added after the original selected-model evaluation; it is not part of that evaluation's original confirmatory endpoint family. Day 10 is a secondary endpoint. We resample complete paired forecasts using 10,000 non-wrapping block-bootstrap draws with seed 20260916 and blocks of five consecutive initialization positions (12 elapsed days). Intervals for reductions, contrasts, and leadwise curves are pointwise 95\% intervals, conditional on the fixed checkpoints and the 2023 sample. They do not measure training-seed or between-year variation, or establish simultaneous significance across leads.

The positive rollout contrast persists with three-start and seven-start blocks, resampling within calendar quarters, and exclusion of the three previously exposed pilot initializations present in this sample (15 February, 15 July, and 15 November 2023, all at 00\,UTC). The latter retains 119 starts and gives $5.27$ percentage points (95\% interval: $4.70$--$5.90$); excluded dates are masked after resampling on the original regular grid. This sensitivity does not remove prior test-year exposure or overlap between forecast windows. Non-wrapping blocks underweight boundary dates, and quarter-wise resampling breaks dependence across quarter boundaries.

Repeated uncorrected GraphCast evaluations differ numerically. For reset/reset, the pooled baseline loss differs by $0.0872\%$ from the common baseline, with a largest initialization/lead difference of $4.39\%$. Using each policy's own repeated baseline instead gives a carry/carry versus reset/reset rollout contrast of $5.34$ percentage points (95\% interval: $4.78$--$5.95$), preserving the conclusion. The reported main estimates consistently use the common baseline.

\subsection{\texorpdfstring{{Mamba capacity and learning-rate selection}}{Mamba capacity and learning-rate selection}}
\label{sec:training_sensitivity}

{Learning-rate choice depends on network width and parameterization. \citet{yang2022tensorprogramsv} derive width-dependent scaling rules under maximal update parameterization ($\mu$P), including an inverse-width scaling for Adam learning rates on hidden weight matrices. In our correction branch, the newly initialized Mamba modules contain approximately 0.11M and 0.22M parameters at inner dimensions 16 and 32, respectively, compared with 10.24M parameters in the spatial components. Motivated by these scaling results and the difference in module size, we test whether the compact Mamba modules benefit from a larger learning rate than the spatial components initialized from pretrained GraphCast weights. We treat this as an empirical hypothesis; the $\mu$P scaling rule does not directly prescribe a learning-rate multiplier for our architecture.}

{We compare Mamba learning-rate multipliers of $0.3\times$, $1\times$, and $3\times$ at each inner dimension, holding the spatial peak learning rate fixed at $10^{-4}$. A multiplier of $1\times$ gives equal learning rates to both parameter groups. All six configurations use spatial width 512, one $B,C$ group, batch size four, and AdamW moments $(0.9,0.98)$. Each run has its own 500-update warm-up followed by 2,000 post-warm-up updates, with SWA over post-warm-up updates 500, 1,000, 1,500, and 2,000. The GraphCast backbone remains frozen throughout. Warm-up parameter values differ across runs despite the common training seed, so the comparisons assess complete training runs.}

\begin{table}[!htbp]
    \centering
    \caption{{Mamba capacity and learning-rate comparison at $1^\circ$. Entries are {GraphCast loss} reductions (\%) on the 32 forecast initializations from the 2022 development set; bold marks the largest value in each metric column. The multiplier applies only to Mamba parameters, with the spatial learning rate fixed. All six configurations use the same training schedule and SWA window.}}
    \label{tab:lr_sensitivity}
    \small
    \begin{tabular}{rrrr}
        \toprule
        $d_{\mathrm{inner}}$ & Mamba LR multiplier & Ten-day aggregate & Day 10 \\
        \midrule
        16 & $0.3\times$ & 14.70 & 21.98 \\
        16 & $1\times$   & 15.67 & 22.44 \\
        16 & $3\times$   & \textbf{16.34} & \textbf{23.68} \\
        32 & $0.3\times$ & 15.82 & 22.96 \\
        32 & $1\times$   & 16.29 & 23.32 \\
        32 & $3\times$   & 15.73 & 22.75 \\
        \bottomrule
    \end{tabular}
\end{table}

\begin{samepage}
Table~\ref{tab:lr_sensitivity} shows that, for inner dimension 16, increasing the Mamba learning-rate multiplier from $1\times$ to $3\times$ raises the ten-day aggregate loss reduction from $15.67\%$ to $16.34\%$ and the day-10 reduction from $22.44\%$ to $23.68\%$. For inner dimension 32, the best tested multiplier is $1\times$; increasing it to $3\times$ lowers the aggregate reduction from $16.29\%$ to $15.73\%$. Dimension 32 performs better at $0.3\times$ and $1\times$, whereas dimension 16 performs better at $3\times$. Thus, the smaller GraphCast-Mamba head benefits from a larger learning rate in these runs, supporting joint selection of Mamba capacity and learning rate. The results do not establish a monotonic benefit from increasing either quantity.\par
\end{samepage}

Across all six tested combinations of Mamba inner dimension and learning-rate multiplier, ten-day aggregate GraphCast loss reductions on the 2022 development set range from $14.70\%$ to $16.34\%$, with day-10 reductions of $21.98$--$23.68\%$. These consistently substantial gains indicate that the overall correction benefit is robust across the tested choices of Mamba capacity and learning rate.

\FloatBarrier
\subsection{\texorpdfstring{{Optimizer configuration study}}{Optimizer configuration study}}
\label{app:optimizer_study}
{These optimizer comparisons use the 2022 development set.}
The optimizer study varies peak learning rate ($10^{-4}$ or $5\times10^{-5}$), Adam moments, Mamba initialization scheme (legacy or Mamba1), and the paired inner-dimension/group settings (16/1 or 64/4). The latter changes two factors simultaneously and does not isolate inner dimension. Runs use 10,000 updates, seed 22, a 200-update warm-up, cosine decay to one tenth of the peak learning rate, and weight decay $10^{-4}$. {The correction branch is trained} with closed-loop feedback, 24-step backpropagation windows, a training tail of 20 feedback-conditioned steps, and loss across all 24 steps. {The forecast correction contribution to the feedback} is stop-gradient, while temporal memory state gradients propagate within the backpropagation window.

Day-10 reductions for the 22 completed selected SWA configurations range from $13.47\%$ to $23.21\%$. The configuration at the upper end uses Mamba1 initialization, inner dimension 16, one $B,C$ group, learning rate $10^{-4}$, and AdamW moments $(0.9,0.98)$; its aggregate loss reduction is $16.28\%$.

\subsection{\texorpdfstring{{Checkpoint averaging and model selection}}{Checkpoint averaging and model selection}}
\label{app:checkpoint_selection}
\begin{samepage}
\paragraph{Stochastic weight averaging.}
{SWA averages correction-branch parameter values} from saved checkpoints to reduce dependence on a single training iterate \citep{izmailov2019averagingweightsleadswider}. Every configuration in {Table~\ref{tab:lr_sensitivity}} averages four checkpoints at {post-warm-up updates} 500, 1,000, 1,500, and 2,000. The dimension-16 Mamba1-initialized optimizer-study configuration described above averages seven checkpoints, saved every 1k updates from 2k through 8k. These are parameter averages, not ensembles of independent forecast predictions. {Evaluation initializes the temporal memory state to zero}, including for SWA models.
\end{samepage}

\paragraph{Training methods and model selection.}
{{Table~\ref{tab:lr_sensitivity} reports all six SWA configurations from the Mamba capacity and learning-rate comparison.} Figure~\ref{fig:res1_forecast} uses the dimension-16, $3\times$ Mamba learning-rate SWA configuration, which has the largest aggregate and day-10 {GraphCast loss} reductions on the 2022 development sample among the completed SWA configurations in the reported optimizer and learning-rate studies.} {This checkpoint is fixed for the 2023 evaluation.}

{{In the optimizer study, each run screens periodic checkpoints and three SWA windows on eight forecast initializations.} The selected checkpoint and SWA are then evaluated on the 32 forecast initializations from the 2022 development set.}

\subsection{Variable-group RMSE}
\label{app:variable_group_rmse}

Table~\ref{tab:variable_group_rmse} provides the numerical results underlying Figure~\ref{fig:breadth_7yr} at selected forecast leads. {It uses the principal seven-year correction and its paired GraphCast baseline on the 122 forecast initializations in 2023, evaluated on the native $1^\circ$ grid as described in Section~\ref{sec:eval_protocol}.}

\begin{table}[!htbp]
    \centering
    \small
    \setlength{\tabcolsep}{4pt}
    \begin{tabular}{@{}lrrrrr@{}}
        \toprule
        & \multicolumn{2}{c}{Day-10 RMSE} & \multicolumn{3}{c}{RMSE reduction (\%)} \\
        \cmidrule(lr){2-3}\cmidrule(l){4-6}
        Variable [unit] & GraphCast & \shortstack{GraphCast--\\Mamba} & Day 5 & Day 7 & Day 10 \\
        \midrule
        \multicolumn{6}{@{}l}{\textit{Upper-air variables (13 pressure levels)}} \\
        Geopotential [m$^2$\,s$^{-2}$] & {784.41} & {703.68} & {1.24} & {5.22} & {10.29} \\
        Temperature [K] & {3.10} & {2.79} & {2.11} & {5.98} & {10.07} \\
        Specific humidity [g\,kg$^{-1}$] & {1.10} & {0.97} & {4.33} & {7.88} & {11.68} \\
        Zonal wind [m\,s$^{-1}$] & {8.67} & {7.63} & {2.76} & {7.19} & {11.94} \\
        Meridional wind [m\,s$^{-1}$] & {9.05} & {7.85} & {2.78} & {7.22} & {13.23} \\
        \shortstack[l]{Pressure vertical velocity\\{}[Pa\,s$^{-1}$]} & {0.177} & {0.164} & {1.89} & {4.52} & {7.09} \\
        \addlinespace
        \multicolumn{6}{@{}l}{\textit{Surface variables}} \\
        $10$\,m zonal wind [m\,s$^{-1}$] & {4.28} & {3.68} & {4.51} & {9.90} & {14.13} \\
        $10$\,m meridional wind [m\,s$^{-1}$] & {4.50} & {3.84} & {4.58} & {9.50} & {14.60} \\
        $2$\,m temperature [K] & {2.67} & {2.41} & {2.66} & {6.26} & {9.59} \\
        Mean sea-level pressure [hPa] & {7.20} & {6.37} & {2.78} & {7.67} & {11.55} \\
        \shortstack[l]{$6$\,h accumulated\\precipitation [mm]} & {2.69} & {2.47} & {2.96} & {5.57} & {8.36} \\
        \bottomrule
    \end{tabular}
    \caption{Physical-unit RMSE and relative RMSE reduction for the principal GraphCast--Mamba model. Days 5, 7, and 10 denote leads of 120, 168, and 240\,h. RMSE pools cosine-latitude-weighted squared errors across forecast initializations and, for upper-air groups, equally across 13 pressure levels before taking the square root. Each reduction is $100(1-\mathrm{RMSE}_{\mathrm{model}}/\mathrm{RMSE}_{\mathrm{GC}})$, computed before rounding. Positive values indicate lower RMSE; these percentages are distinct from GraphCast loss reductions.}
    \label{tab:variable_group_rmse}
\end{table}

\clearpage
\subsection{Computational resources}
\label{app:computational_resources}

Table~\ref{tab:computational_resources} reports the measured resources for the selected model. The elapsed time covers the complete GPU allocation for the 500-update initialization stage, the 2,000-update main stage, compilation, data loading, checkpointing, and SWA construction. Data preparation, hyperparameter search, and forecast evaluation are excluded. Forecast evaluation ran separately, and neither training stage performed inline validation. Host memory is the peak resident memory reported by Slurm; GPU memory is the largest value observed in five-second telemetry samples.

\begin{table}[!htbp]
    \centering
    \small
    \begin{tabular}{@{}lr@{}}
        \toprule
        Quantity & Selected model \\
        \midrule
        Accelerator & 1 NVIDIA A100 80GB PCIe \\
        Allocated CPU cores & 8 \\
        Elapsed allocation time & 22\,h 40\,min 45\,s \\
        GPU allocation cost & 22.68 GPU-hours \\
        Peak host memory & 97.1 GiB \\
        Sampled peak GPU memory & 23.15 GiB \\
        \addlinespace
        Frozen backbone parameters & 35,979,347 \\
        Trainable correction parameters & 10,348,879 \\
        \quad Spatial components and output head & 10,238,863 \\
        \quad Mamba modules & 110,016 \\
        \bottomrule
    \end{tabular}
    \caption{Training resources and parameter counts for the selected correction model. GPU-hours equal elapsed allocation hours because the run uses one GPU. The GPU-memory value is a sampled peak.}
    \label{tab:computational_resources}
\end{table}

\clearpage
\section{Training tail length}
\label{app:training_horizon}
{These archived training-tail sweeps use 2022 forecasts and separate training configurations.}

We examine how forecast skill depends on the training tail length $K$ under two input-feedback policies. Each 24-step training window begins with a \emph{teacher-forced prefix}: the model receives ERA5 states as meteorological inputs while updating and retaining its temporal memory. For $K>1$, the final $K$ steps instead use forecast feedback, leaving $24-K$ teacher-forced steps. The label $K=1$ denotes fully teacher-forced training, with ERA5 inputs at every step. We evaluate the resulting models over 40 six-hour forecast steps (ten days) and report {GraphCast loss} reduction relative to GraphCast; positive values indicate improved forecast skill.

\subsection{{Closed-loop training}}
\begin{figure}[!htbp]
    \centering
    \includegraphics[width=\linewidth]{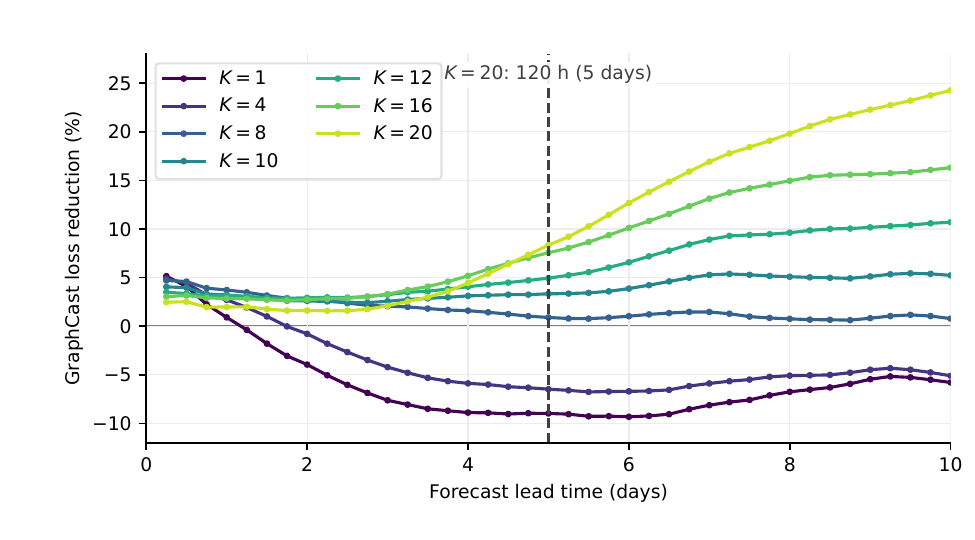}
    \caption{Forecast skill after closed-loop training with different training tail lengths $K$. During the tail, corrected GraphCast--Mamba predictions supply subsequent inputs. Each curve shows {GraphCast loss} reduction relative to GraphCast over ten forecast days; positive values indicate improvement. {The vertical dashed line marks $20\times6=120$\,h (5 days), the training tail duration for $K=20$. Gains for this model persist beyond its training tail to day 10, twice that duration.}}
    \label{fig:training_horizon_closed_loop}
\end{figure}

In \emph{closed-loop training}, the corrected GraphCast--Mamba forecast supplies the next meteorological input during the training tail. The learned correction therefore affects both the current prediction and the inputs encountered at subsequent steps. Figure~\ref{fig:training_horizon_closed_loop} compares models trained with different $K$, showing how the duration of this feedback changes forecast skill.

Short training tails primarily improve early forecasts. Models trained with $K=1$ and $K=4$ reduce error during the first one to two days but increase it thereafter, whereas longer training tails yield substantial gains at longer lead times. At day 10, the error reduction relative to GraphCast increases from 5.2\% for $K=10$ to 24.3\% for $K=20$. These curves reveal a crossover: shorter tails favor early forecast skill, while longer tails become more effective as the forecast advances. To examine this crossover in a more controlled setting, we remove correction feedback and train the Mamba correction branch on precomputed forecast trajectories from the frozen GraphCast backbone. In this open-loop setting, corrections do not affect subsequent backbone predictions. As shown in the next section, the crossover becomes more pronounced under this training scheme.

\clearpage
\subsection{Open-loop training and a sharper crossover}
In \emph{open-loop training}, the teacher-forced prefix is followed by a tail driven by the uncorrected GraphCast forecasts. The correction branch learns along this trajectory, but its predictions do not alter subsequent meteorological inputs. Thus, open loop refers to the absence of correction feedback, while teacher forcing refers to the use of ERA5 inputs in the prefix.

\begin{figure}[!htbp]
    \centering
    \includegraphics[width=\linewidth]{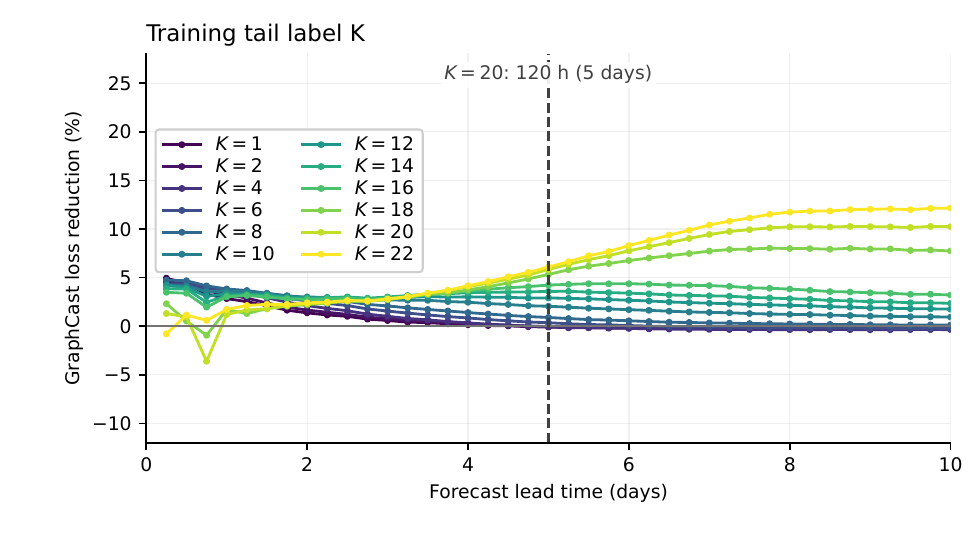}
    \caption{Forecast skill after open-loop training with a teacher-forced prefix and different training tail lengths $K$. During the tail, uncorrected GraphCast forecasts supply subsequent inputs. {GraphCast loss} reduction exhibits a pronounced crossover, with substantially larger long-lead gains for $K\geq18$. {The vertical dashed line marks $20\times6=120$\,h (5 days), the training tail duration for $K=20$; this model retains positive gains beyond that duration, up to day 10.} This figure and Figure~\ref{fig:training_horizon_closed_loop} use the same vertical scale but compare separate sweeps with different experimental settings and loss reconstructions (Appendix~\ref{app:training_loss}).}
    \label{fig:training_horizon_open_loop}
\end{figure}

In our experiments, training with this teacher-forced prefix and a tail independent of the learned correction was more stable but {yielded smaller improvements at longer lead times}. Figure~\ref{fig:training_horizon_open_loop} shows that short-tail models remain close to GraphCast at long lead times, avoiding the substantial degradation seen in the closed-loop comparison. The crossover is more pronounced here: long-lead gains change markedly between $K=16$ and $K=18$ and increase further for $K=20$ and $K=22$.

This crossover raises an interesting question about how temporal memory, training tail length, and feedback interact. Its mechanism, including why it is sharper under open-loop training, remains unclear. We regard it as an empirical finding that motivates further investigation, rather than evidence for a universal critical value of $K$.

\clearpage

\section{Comparison on a common verification grid}
\label{app:common_grid}
\begin{figure}[!t]
    \centering
    \includegraphics[width=\linewidth]{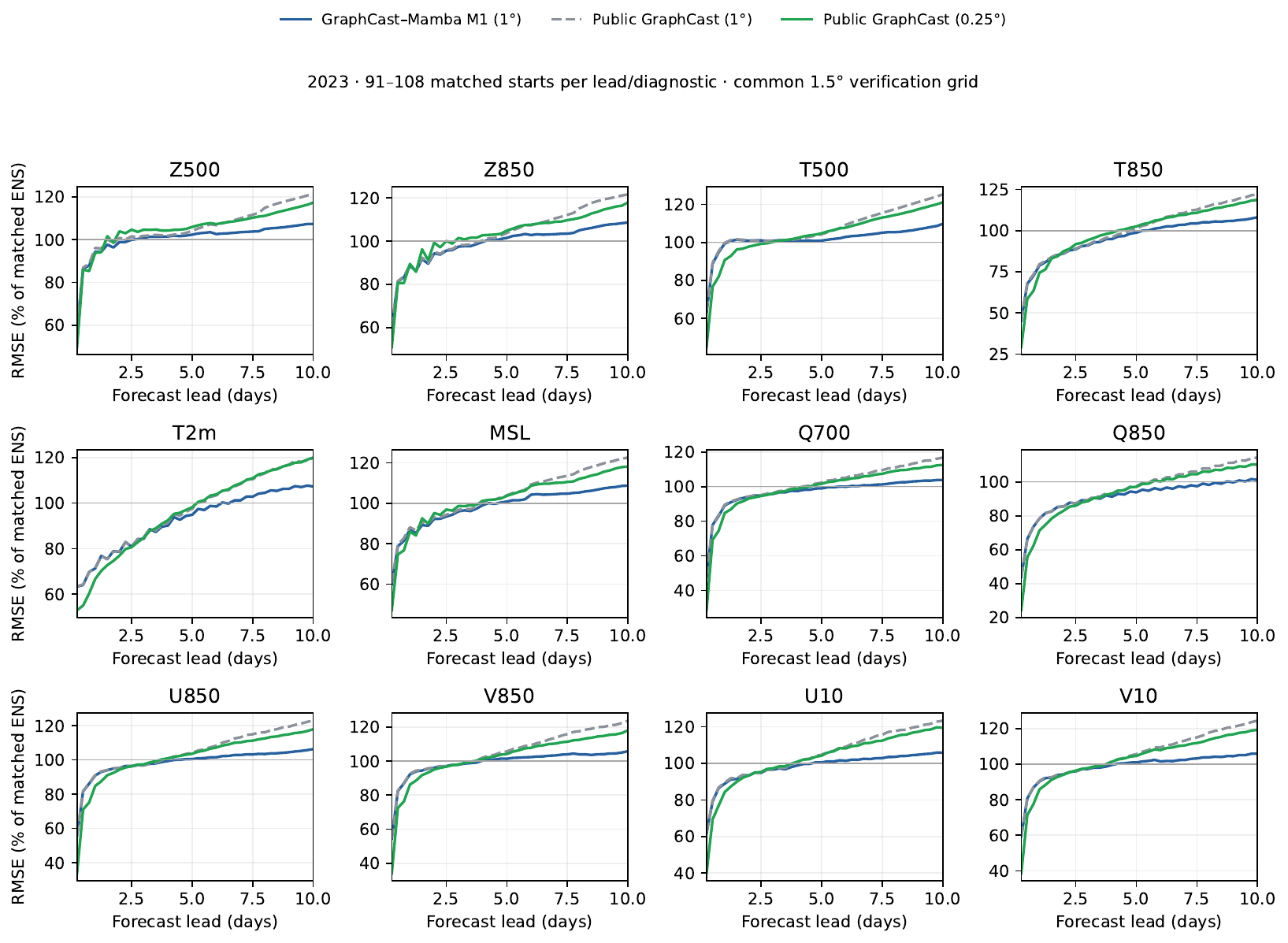}
    \caption{The comparison in Figure~\ref{fig:common_grid_selected} extended to 12 diagnostics. Blue shows the selected GraphCast--Mamba model M1 ($1^\circ$), dashed gray its public GraphCast\_small backbone ($1^\circ$), and green the public GraphCast checkpoint ($0.25^\circ$). Z denotes geopotential, T temperature, Q specific humidity, U zonal wind, and V meridional wind; a three-digit suffix gives pressure in hPa. T2m denotes $2$\,m temperature, U10 and V10 the $10$\,m wind components, and MSL mean sea-level pressure. All four forecasts use identical dates and ERA5 truth at each diagnostic and lead on the $1.5^\circ$ grid; the available date set varies across points (91--108 starts). Curves show pooled RMSE as a percentage of the matched ENS-mean RMSE, with ENS at $100\%$.}
    \label{fig:res1_all_variables}
\end{figure}

\paragraph{Models and forecast protocol.}
All three GraphCast-based forecasts are generated locally from fixed checkpoints. M1 is the same selected correction used in the principal evaluation, trained on 2015--2021 and selected on 2022 development data. Its frozen backbone is the public DeepMind GraphCast\_small checkpoint trained on ERA5 1979--2015, with a $1^\circ$ grid, 13 pressure levels and mesh refinement level 5. The higher-resolution reference is the public GraphCast checkpoint trained on ERA5 1979--2017, with a $0.25^\circ$ grid, 37 levels and mesh refinement level 6. Both public checkpoints include precipitation as an input and output and use the supplied GraphCast normalization statistics. Their grids contain $181\times360$ and $721\times1440$ points, respectively. Their meshes contain 10,242 and 40,962 vertices, respectively. The higher-resolution model therefore operates on much larger spatial graphs; the computational comparison concerns FLOPs and activation memory. Exact checkpoint filenames, checksums and configuration are recorded in the figure-source manifest.

Each forecast receives observed ERA5 states at $-6$ and $0$\,h, then advances through 40 six-hour leads from 6 to 240\,h using its own forecast feedback. Time-varying $0.25^\circ$ ERA5 inputs are subsampled to $1^\circ$ for GraphCast\_small and M1 and restricted to their 13 pressure levels; these models retain static fields from their existing training and evaluation store. M1 begins each forecast with zero temporal memory and feeds back its full corrected prediction. Network inference uses bfloat16, with physical fields and state boundaries in float32.

\paragraph{Dates and ENS reference.}
The candidate sample consists of 122 initializations at 00\,UTC every third day from 1 January through 30 December 2023, with verification extending through 9 January 2024. The evaluation protocol records earlier numerical-pilot exposure, including three candidate starts (15 February, 15 July and 15 November); no checkpoint selection used this comparison. ENS denotes the ECMWF IFS ensemble-mean forecast distributed by WeatherBench~2 on its conservative $121\times240$ grid. We use available 2023 forecasts from the extended 2016--2024 store.\footnote{\raggedright The ENS stores under \url{gs://weatherbench2/datasets/ifs_ens/} are \path{2016-2024-240x121_equiangular_with_poles_conservative_mean.zarr} (forecast source) and \path{2018-2022-240x121_equiangular_with_poles_conservative_mean.zarr} (layout reference).} Missing metadata and coordinate arrays were reconstructed from the complete 2018--2022 sibling store, with decoding checked against 71 overlapping forecast chunks. Missing forecast chunks remain unavailable.

At each diagnostic and lead, we retain exactly the same ENS-available initializations for all four forecasts. This gives 91--108 starts per point across the 12 diagnostics (92--105 in Figure~\ref{fig:common_grid_selected}), with 46,930 initialization--lead--diagnostic combinations per model. The matched date set can vary between diagnostics and leads; there is no single complete subset underlying every point. Selection depends only on forecast availability, not errors. Availability masks and the full candidate manifest accompany the figures.

\paragraph{Verification and metric.}
Every forecast is verified against identical ERA5 fields~\citep{hersbach2020era5} from ARCO ERA5, with source-data repairs validated against the independently curated NSF NCAR ERA5 archive. Native model predictions and $0.25^\circ$ verification fields are remapped to the common $1.5^\circ$ grid using first-order conservative spherical-cell overlap with periodic longitude; the archived ENS fields already occupy this grid. For each diagnostic and lead, we compute spatial mean squared errors with normalized cosine-latitude weights, average equally over the matched starts, and take the square root. The plotted value is $100\,\mathrm{RMSE}_{\mathrm{model}}/\mathrm{RMSE}_{\mathrm{ENS}}$, with identical dates in numerator and denominator. These physical-unit RMSE diagnostics are separate from the normalized GraphCast training loss.

\paragraph{Interpretation.}
The figures show point estimates for the specified fixed checkpoints on the available matched samples. Ten-day verification windows overlap, so the starts are not independent weather realizations. Checkpoints also differ in architecture, pressure levels and training coverage; their comparison does not isolate resolution or attribute the full advantage to memory. The matched memory controls in Section~\ref{sec:memory_ablation} address the latter question. Lower RMSE after common-grid remapping establishes neither finer-scale skill at $0.25^\circ$ nor a measured inference speedup. At day 10, M1 has lower RMSE than the public $0.25^\circ$ checkpoint on all 12 diagnostics, but higher RMSE than ENS.

\clearpage
\section{Forecast examples at city locations}
\label{app:city_traces}
\begin{figure}[!htbp]
    \centering
    \includegraphics[width=\linewidth]{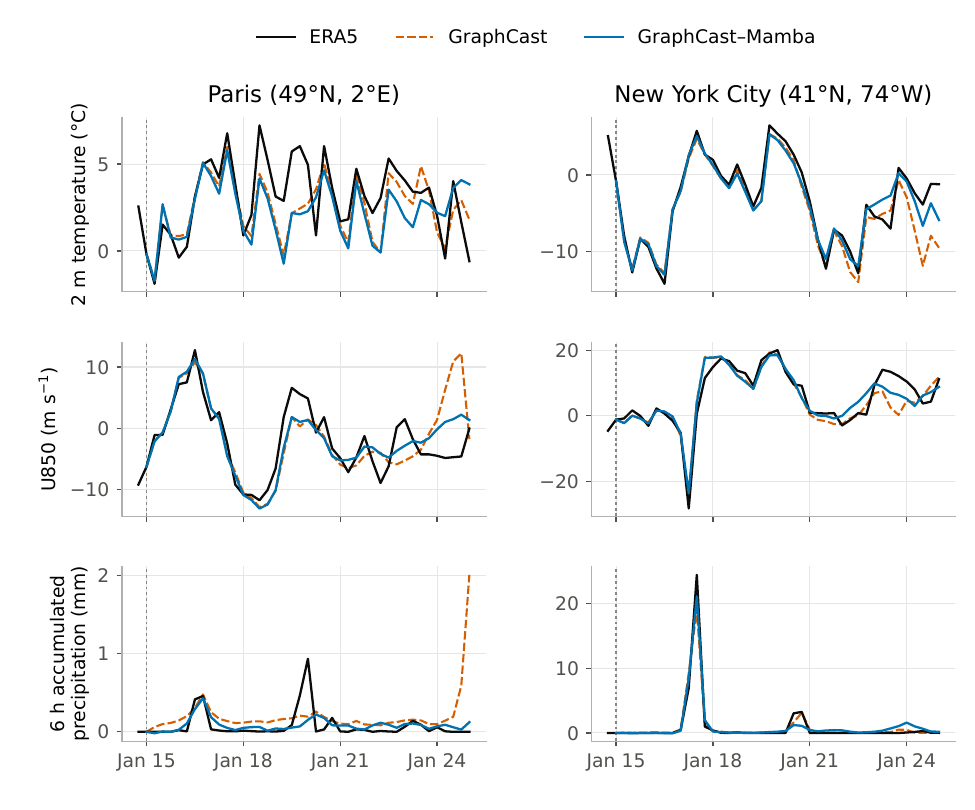}
    \caption{{Ten-day forecasts from 00\,UTC on 15 January 2022 at the $1^\circ$ grid points nearest Paris ($49^\circ$N, $2^\circ$E; left) and New York City ($41^\circ$N, $74^\circ$W; right).
    From top to bottom, rows show $2$\,m temperature (T2m), $850$\,hPa zonal wind (U850), and $6$\,h accumulated precipitation in physical units.
    ERA5 is shown in black, the frozen GraphCast forecast in dashed orange, and the GraphCast--Mamba forecast in blue.
    The correction was trained separately on 15 years of ERA5 data (2007--2021), with a training tail of $K=20$ feedback-conditioned steps, Mamba inner dimension $128$, spatial width $512$, and two graph message-passing steps; SWA averages checkpoints from updates 6k--11k.
    Both forecasts use the ERA5 states at $-6$ and $0$\,h (before and on the dotted line) as inputs, then feed back their own predicted fields at each step; the GraphCast--Mamba model begins with a zero temporal memory state.
    Precipitation is shown as predicted and may take small negative values.}}
    \label{fig:city_traces_jan}
\end{figure}

To complement the aggregate seven-year results in Section~\ref{sec:numerical}, Figure~\ref{fig:city_traces_jan} illustrates local forecast behavior for temperature, wind, and precipitation at Paris and New York City. These examples use a separate correction trained on 15 years of ERA5 data (2007--2021), with the configuration specified in the figure caption. Over 120--240\,h, the correction reduces mean absolute error (MAE) in four of the six displayed city--variable pairs. Errors increase for Paris temperature and New York precipitation, illustrating that aggregate skill gains can coexist with deterioration in individual forecasts.

\end{document}